%% file: main.tex
\documentclass{article}

\input{includes/base.tex}

\input{includes/commands.tex}
\input{includes/symbols.tex}

\title{HPTQ: Hardware-Friendly Post Training Quantization}
\author{Hai Victor Habi, Reuven Peretz, Elad Cohen, Lior Dikstein, \\ Oranit Dror, Idit Diamant, Roy H. Jennings and Arnon Netzer\\
Sony Semiconductor Israel\\
{\tt\small hai.habi@sony.com}\\
}

\date{July 2021}

\begin{document}

\maketitle
\begin{abstract}
     Neural network quantization enables the deployment of models on edge devices. 
     An essential requirement for their hardware efficiency is that the quantizers are hardware-friendly: uniform, symmetric and with power-of-two thresholds. 
     To the best of our knowledge, current post-training quantization methods do not support all of these constraints simultaneously.
     In this work we introduce a hardware-friendly post training quantization (HPTQ) framework, which addresses this problem by synergistically combining several known quantization methods.
    %  , namely, threshold selection, shift negative correction, channel equalization, per channel quantization and bias correction. 
     We perform a large-scale study on four tasks: classification, object detection, semantic segmentation and pose estimation over a wide variety of network architectures.
     Our extensive experiments show that competitive results can be obtained under hardware-friendly constraints.
\end{abstract}

%as well as original modifications to existing methods that are used in unconstrained quantization and modified/applied them on the constrained quantization scheme

     %In addition, we overview a set of methods that are used in unconstrained quantization and modified/applied them on the constrained quantization scheme. 

\input{files/introduction}
\input{files/background}
\input{files/bag_of_tricks}
%\input{files/kd_cptq}
\input{files/experimental_results}
\section{Conclusions}\label{sec:conculsions}
In this work we propose HPTQ, a method for hardware-friendly post-training quantization. 
HPTQ offers a flow that adapts and synergistically combines several known quantization techniques both for weights and activations.
%, namely, threshold selection, shift negative correction, channel equalization, per channel quantization and bias correction.
We extensively evaluated the performance of HPTQ on four tasks: classification, object detection, semantic segmentation and pose estimation. 
% For all of the tasks we obtain competitive results for hardware friendly quantization: uniform, symmetric quantization with power-of-two thresholds.
Notably, for all of the tasks we demonstrated that competitive results can be obtained under our hardware-friendly constraints of uniform and symmetric quantization with power-of-two thresholds.
% Our experiments have shown that HPTQ has the advantage of utilizing constraint quantizers for low computational cost with only a slight degradation in performance. 
In addition, we performed an ablation study in which we presented the contributions of each of the methods used by HPTQ.

\bibliographystyle{unsrt}
\bibliography{ref}

\end{document}

%% file: includes/base.tex
\usepackage[utf8]{inputenc}
\usepackage{amssymb}
\usepackage{bm}
\usepackage[ruled,vlined]{algorithm2e}
\usepackage{hyperref}

\usepackage{graphicx}
\usepackage{float}
\usepackage{amsmath}
\usepackage{tikz}
\usepackage{lmodern}
\usepackage{subcaption}
\usepackage{bbm}
\usepackage{multirow}
\usetikzlibrary{positioning, backgrounds, fit, shapes.arrows}
\usetikzlibrary{decorations.markings}
\usetikzlibrary{shapes.geometric,shapes.symbols,fit,positioning,shadows}

\usepackage{amsthm}

\makeatletter
\tikzstyle{dummy} = [coordinate]
\pgfkeys{/pgf/.cd,
  parallelepiped offset x/.initial=2mm,
  parallelepiped offset y/.initial=2mm
}
\pgfdeclareshape{parallelepiped}
{
  \inheritsavedanchors[from=rectangle] % this is nearly a rectangle
  \inheritanchorborder[from=rectangle]
  \inheritanchor[from=rectangle]{north}
  \inheritanchor[from=rectangle]{north west}
  \inheritanchor[from=rectangle]{north east}
  \inheritanchor[from=rectangle]{center}
  \inheritanchor[from=rectangle]{west}
  \inheritanchor[from=rectangle]{east}
  \inheritanchor[from=rectangle]{mid}
  \inheritanchor[from=rectangle]{mid west}
  \inheritanchor[from=rectangle]{mid east}
  \inheritanchor[from=rectangle]{base}
  \inheritanchor[from=rectangle]{base west}
  \inheritanchor[from=rectangle]{base east}
  \inheritanchor[from=rectangle]{south}
  \inheritanchor[from=rectangle]{south west}
  \inheritanchor[from=rectangle]{south east}
  \backgroundpath{
    \southwest \pgf@xa=\pgf@x \pgf@ya=\pgf@y
    \northeast \pgf@xb=\pgf@x \pgf@yb=\pgf@y
    \pgfmathsetlength\pgfutil@tempdima{\pgfkeysvalueof{/pgf/parallelepiped offset x}}
    \pgfmathsetlength\pgfutil@tempdimb{\pgfkeysvalueof{/pgf/parallelepiped offset y}}
    \def\ppd@offset{\pgfpoint{\pgfutil@tempdima}{\pgfutil@tempdimb}}
    \pgfpathmoveto{\pgfqpoint{\pgf@xa}{\pgf@ya}}
    \pgfpathlineto{\pgfqpoint{\pgf@xb}{\pgf@ya}}
    \pgfpathlineto{\pgfqpoint{\pgf@xb}{\pgf@yb}}
    \pgfpathlineto{\pgfqpoint{\pgf@xa}{\pgf@yb}}
    \pgfpathclose
    \pgfpathmoveto{\pgfqpoint{\pgf@xb}{\pgf@ya}}
    \pgfpathlineto{\pgfpointadd{\pgfpoint{\pgf@xb}{\pgf@ya}}{\ppd@offset}}
    \pgfpathlineto{\pgfpointadd{\pgfpoint{\pgf@xb}{\pgf@yb}}{\ppd@offset}}
    \pgfpathlineto{\pgfpointadd{\pgfpoint{\pgf@xa}{\pgf@yb}}{\ppd@offset}}
    \pgfpathlineto{\pgfqpoint{\pgf@xa}{\pgf@yb}}
    \pgfpathmoveto{\pgfqpoint{\pgf@xb}{\pgf@yb}}
    \pgfpathlineto{\pgfpointadd{\pgfpoint{\pgf@xb}{\pgf@yb}}{\ppd@offset}}
  }
}
\pgfdeclareshape{document}{
\inheritsavedanchors[from=rectangle] % this is nearly a rectangle
\inheritanchorborder[from=rectangle]
\inheritanchor[from=rectangle]{center}
\inheritanchor[from=rectangle]{north}
\inheritanchor[from=rectangle]{north east}
\inheritanchor[from=rectangle]{north west}
\inheritanchor[from=rectangle]{south}
\inheritanchor[from=rectangle]{south east}
\inheritanchor[from=rectangle]{south west}
\inheritanchor[from=rectangle]{west}
\inheritanchor[from=rectangle]{east}
\backgroundpath{%
\southwest \pgf@xa=\pgf@x \pgf@ya=\pgf@y
\northeast \pgf@xb=\pgf@x \pgf@yb=\pgf@y
\pgf@xc=\pgf@xb \advance\pgf@xc by-5pt % this should be a parameter
\pgf@yc=\pgf@ya \advance\pgf@yc by5pt
\pgfpathmoveto{\pgfpoint{\pgf@xa}{\pgf@ya}}
\pgfpathlineto{\pgfpoint{\pgf@xa}{\pgf@yb}}
\pgfpathlineto{\pgfpoint{\pgf@xb}{\pgf@yb}}
\pgfpathlineto{\pgfpoint{\pgf@xb}{\pgf@yc}}
\pgfpathlineto{\pgfpoint{\pgf@xc}{\pgf@ya}}
\pgfpathclose
\pgfpathmoveto{\pgfpoint{\pgf@xc}{\pgf@ya}}
\pgfpathlineto{\pgfpoint{\pgf@xc}{\pgf@yc}}
\pgfpathlineto{\pgfpoint{\pgf@xb}{\pgf@yc}}
\pgfpathclose
}
}
\tikzstyle{block} = [draw, fill=white, rectangle, 
    minimum height=3em, minimum width=6em]
    
\usetikzlibrary{backgrounds}
\usepackage{pifont}% http://ctan.org/pkg/pifont
\tikzstyle{startstop} = [rectangle, rounded corners, minimum width=2cm, minimum height=1cm,text centered, draw=black, fill=lime!30]

%% file: includes/commands.tex
\newcommand{\vectorsym}[1]{\bm{#1}}
\newcommand{\expectation}[2]{\mathbb{E}_{#2}\left[#1\right]}

\newcommand{\brackets}[1]{\left(#1\right)}
\newcommand{\abs}[1]{\left|#1\right|}

\newcommand{\round}[1]{\left\lfloor#1\right\rceil}

\newcommand{\matsym}[1]{\mathbf{#1}}

\newcommand{\clip}[3]{\mathrm{clip}\left(#1,#2,#3\right)}
\newcommand{\ceil}[1]{\left\lceil#1\right\rceil}

\newcommand{\xmark}[0]{\ding{55}}

%% file: includes/symbols.tex
\newcommand{\mbvone}{MobileNetV1 \cite{howard2017mobilenets} }
\newcommand{\mbvtwo}{MobileNetV2 \cite{sandler2018mobilenetv2} }
\newcommand{\nasnet}{NasnetMobile \cite{zoph2018learning} }
\newcommand{\vgg}{VGG16      \cite{simonyan2014very} }
\newcommand{\inc}{InceptionV3 \cite{szegedy2016rethinking} }
\newcommand{\incres}{InceptionResNetV2 \cite{szegedy2017inception} }
\newcommand{\res}{ResNet50 \cite{he2016deep} }
\newcommand{\eff}{EfficientNet-B0 \cite{tan2019efficientnet} }
\newcommand{\effrelu}{EfficientNet-B0 ReLU$^{\ref{foot:eff}}$}

\newcommand{\dense}{DenseNet-121 \cite{huang2017densely} }
\newcommand{\xecption}{Xception \cite{chollet2017xception} }
\newcommand{\cptq}{HPTQ (Our) }
\newcommand{\tqt}{TQT \cite{jain2019trained}}
\newcommand{\ssbd}{SSBD \cite{meller2019same}}
\newcommand{\lee}{Lee et al \cite{lee2018quantization}}
\newcommand{\qt}{QT \cite{jacob2018quantization}}
\newcommand{\wu}{Wu et al \cite{wu2020integer}}

\newcommand{\nagel}{Nagel et al \cite{nagel2021white}}
\newcommand{\zeroq}{ZeroQ \cite{cai2020zeroq}}
\newcommand{\dfq}{DFQ \cite{nagel2019data}}
\newcommand{\adaquant}{AdaQuant \cite{hubara2020improving}}
\newcommand{\Krishnamoorthi}{Krishnamoorthi \cite{krishnamoorthi2018quantizing}}
\newcommand{\rvquant}{RVQuant \cite{park2018value}}
\newcommand{\hawq}{HAWQ-V3 \cite{yao2021hawq}}
\newcommand{\ocs}{OCS \cite{zhao2019improving}}
\newcommand{\faq}{FAQ \cite{mckinstry2019discovering}}
\newcommand{\lsq}{LSQ \cite{esser2019learned}}
\newcommand{\he}{He et al \cite{he2018learning}}

%% file: files/introduction.tex
\section{Introduction}
% Motivation for quantization
{\let\thefootnote\relax\footnote{{An implementation of this work is available at \url{https://github.com/sony/model_optimization}.}}}
Deep neural networks have shown state-of-art performance in many real-world computer vision tasks, such as image classification \cite{he2016deep,howard2017mobilenets}, object detection \cite{ren2015faster,liu2016ssd,lin2017feature}, semantic segmentation \cite{chen2017rethinking} and pose estimation \cite{zhang2019simple,cao2019openpose}. However, the deployment of deep neural networks on edge devices is still considered a challenging task due to limitations on available memory, computational power and power consumption.

Quantization \cite{gholami2021survey} is a common approach to tackle this challenge with minimal performance loss, by reducing the bit-width of network weights and activations.
%to overcome device limitations. 
Quantization methods can be roughly divided into two categories: quantization aware training (QAT) and post-training quantization (PTQ).  
QAT methods \cite{jacob2018quantization,jain2019trained,choi2018pact,gong2019differentiable} retrain the network in order to recover the accuracy degradation caused by quantization and usually achieve better results than PTQ methods. 
% However, these methods require task-specific knowledge for the retraining process, such as labeled data, loss function and data augmentation. 
PTQ methods \cite{banner2018post,cai2020zeroq,nagel2020up,fang2020post} are simpler and add quantization to a given network model without any training process. 
These methods are usually based on a representative unlabeled dataset that is used for selecting the quantization parameters. 

Recently, several works \cite{jain2019trained,hmq,uhlich2019mixed} have focused on \textit{hardware friendly quantization} schemes.
Namely, that their quantizers are uniform, symmetric and with power-of-two thresholds. 
Such quantizers optimize computational costs as they allow integer arithmetic without any cross-terms due to zero-points and floating-point scaling \cite{jain2019trained}.
% A detailed analysis of the computational benefits of using hardware friendly quantization is shown in Appendix \ref{sec:aconstrained}. 

% A detailed analysis of the computational overhead of unconstrained quantization compared with constrained quantization is shown in Appendix \ref{sec:aconstrained}. 
%The analysis includes an evaluation of the total overhead associated with several network architectures for different parameters such as width, resolution and depth. 

In this work, we introduce a hardware-friendly post-training quantization (HPTQ) method.
To the best of our knowledge, current hardware friendly quantization methods are based on quantization aware training (QAT). 
This might be due to the difficulty of using power-of-two thresholds as stated in \cite{nagel2021white}. 
HPTQ offers a post-training quantization flow that adapts and synergistically combines several known techniques, namely, threshold selection, shift negative correction, channel equalization, per channel quantization and bias correction. 

We extensively examine the performance of our method using 8-bit quantization. 
We evaluate HPTQ on different network architectures over a variety of tasks, including classification, object detection, semantic segmentation and pose estimation. Additionally, we provide an ablation study demonstrating the effect of each technique on the network performance.
To summarize, our contributions are:
\begin{itemize}
    \item Introducing HPTQ, a method for hardware friendly post-training quantization.
    \item A large-scale study of post-training quantization on a variety of tasks: classification, object detection, semantic segmentation and pose estimation.
    \item We demonstrate that competitive results can be obtained under hardware friendly constraints of uniform, symmetric 8-bit quantization with power-of-two thresholds.
\end{itemize}

%In this work we give a comprehensive analysis of a wide variety of known quantization tricks that can be employed or adapted for a "hardware friendly" quantizer. 

%Based on this analysis, we introduce a constrained post-training quantization (CPTQ) method that combines all these tricks as well as new ones into a single framework for constrained quantization. 

%% file: files/background.tex
\section{Background and Basic Notions}
\label{sec:background}

In this section we give a short overview of uniform quantization and the hardware friendly constraints that will be applied in this work, namely, symmetric quantization with power-of-two thresholds.

\paragraph{Uniform Affine Quantization.}

% A uniform quantizer $Q$ is a real valued step function that maps real numbers to a discrete set of quantized numbers that are uniformly spaced. 

A quantizer can be formalized as a right to left composition $Q=Q^{de} \circ Q^{int}$ of an integer valued function $Q^{int}: \mathbb{R} \rightarrow \mathbb{Z}$ and a recovering affine operation $Q^{de}: \mathbb{Z} \rightarrow\mathbb{R}$ (known as \textit{de-quantization}). The discrete range of $Q$ is called a \textit{quantization grid} and if it is uniformly spaced, then $Q$ is said to be a \textit{uniform quantizer}. 

The constant gap between two adjacent points in the quantization grid of a uniform quantizer is called its \textit{step size} and the affine shift is called the \textit{zero point} $z$. 
Using these parameters, a uniform quantizer can be formalized as:

\begin{equation} 
    Q(x)=Q^{de}(Q^{int}(x))=s \cdot \brackets{x^{int}+z}  \approx x
 \label{eq:q_de}
\end{equation}
 where $x_{int}$ is the image of $Q^{int}(x)$ and is called the \textit{quantized integer value} of $x$.
 
Practically, $Q^{int}$ is defined by a \textit{clipping range} of real values $[a,b] \subseteq\mathbb{R}$ and the number of bits $n_b\in\mathbb{N}$ for representing the quantized integer values:

\begin{equation}
x^{int}=Q^{int}\brackets{x,a,b,n_b}=\round{\frac{\clip{x}{a}{b}-a}{s}}
\end{equation}
where $s=\frac{b-a}{2^{n_b}-1}$ is the step size, $\clip{x}{a}{b} = \min(\max(x,a), b)$ and $\round{\cdot}$ is the rounding function to the nearest integer. The zero-point is then defined as $z=\frac{a}{s}$ and a uniform quantizer can be formalized as:
\begin{equation}
    Q\brackets{x,a,b,n_b}=Q^{de}\brackets{Q^{int}\brackets{x,a,b,n_b}}=s\round{\frac{\clip{x}{a}{b}-a}{s}}+a
\label{eq:uniform_quantizer}    \end{equation}

% Finally, the composition of the two functions above provides the following definition of uniform quantization:

Note that usually the clipping boundaries $a,b$ are selected so that the real value 0.0 is a point on the quantization grid. 
%This ensures that real zero is quantized without an error, which might be caused by either the clipping or rounding operation in Eq. \ref{eq:uniform_quantizer}. 

% Note that the difference between $x$ and its quantized counterpart $x^q$ is called its \textit{quantization noise}. 

% The process of defining a uniform quantizer involves the selection of a \textit{clipping range} of real values $(a,b) \subseteq\mathbb{R}$ and the number of bits $n_b\in\mathbb{N}$ for representing the quantized integer values. 

\paragraph{Symmetric Quantization.} 

Symmetric quantization is a simplified case of a uniform quantizer that restricts the zero-point to $0$. This eliminates the need for zero-point shift in Eq. \ref{eq:q_de} and thus enables efficient hardware implementation of integer arithmetic without any cross-terms \nolinebreak \cite{jain2019trained}.

The zero-point restriction to 0 requires the selection of either a signed or unsigned quantization grid. Let $t \in \mathbb{R^{+}}$ be a clipping threshold of the quantization range. 
A \textit{signed quantizer} is then formalized as: 

\begin{equation}
\label{equ:signed_quantizer}
\matsym{x}_{int}=\clip{\round{\frac{\matsym{x}}{s}}}{-2^{n_b-1}}{2^{n_b-1}-1}
\end{equation}

%\begin{equation}
%\label{equ:signed_quantizer}
%Q^{\rm{s}}(\matsym{x}, t,n_b)=s\brackets{t,n_b}\cdot\clip{\round{\frac{\matsym{x}}{s\brackets{t,n_b}}}+2^{n_b-1}}{0}{2^{n_b}-1}-2^{n_b-1}
%\end{equation}
where $s = \frac{2t}{2^{n_b}}$ is the step-size. Similarly, an \textit{unsigned quantizer} is formalized as: 
\begin{equation}
\label{equ:unsigned_quantizer}
\matsym{x}_{int}=\clip{\round{\frac{\matsym{x}}{s}}}{0}{2^{n_b}-1}
\end{equation}
where $s =\frac{t}{2^{n_b}}$ is the step size. 

% Note that throughout this work, whenever a quantizer is applied to an input other than a real value (e.g. a tensor or a vector) it is understood that it is applied element-wise and returns a quantized object with the same shape.

\paragraph{Power-of-Two Thresholds.}
A uniform, symmetric quantizer (either signed or unsigned) with a power-of-two integer threshold is said to be a \textit{hardware-friendly quantizer} \cite{hmq}. Restricting the threshold of a symmetric quantizer to power-of-two integers (i.e. $t=2^M$, where $M \in \mathbb{Z}$) enables an efficient hardware implementation that uses integer arithmetic without floating-point scaling \cite{jain2019trained}.
% Similarly to \cite{hmq}, we name a symmetric quantizer (either signed or unsigned) with a threshold $t=2^M$, where $M \in \mathbb{Z}$, as a \textit{hardware-friendly quantizer}.

% The definition of the quantizer that we use in this work is similar to the one in \cite{jain2019trained,hmq}. The quantizer is parameterized by a pair ($t$, $n_b$) of a threshold $t=2^M$ and a bit-width
% $n_b$, where $M \in \mathbb{Z}$. The signed version $Q^{\rm{s}}$ of a quantizer is defined in Eq. \ref{equ:signed_quantizer} and the unsigned version $Q^{\rm{us}}$ defined in Eq. \ref{equ:unsigned_quantizer}. 

\begin{figure}[H]
    \centering
    \includegraphics[scale=0.35]{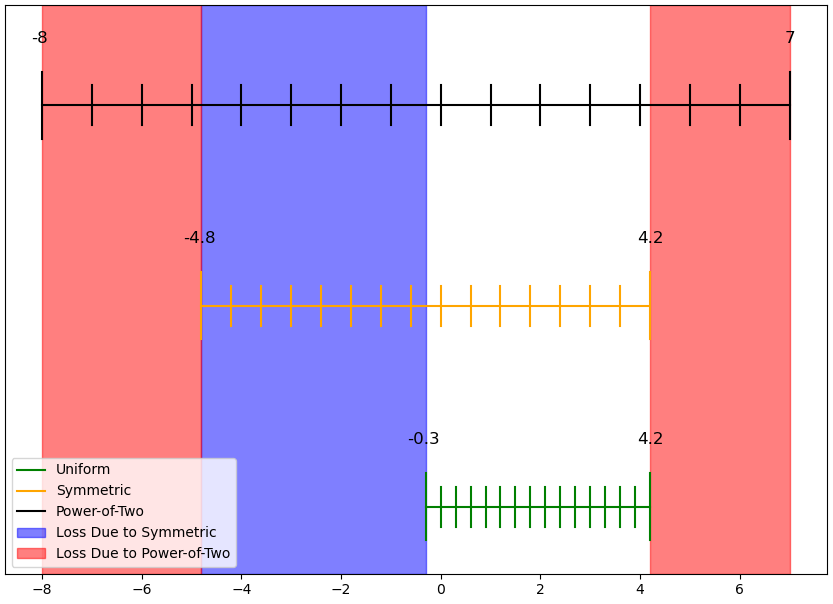}
    \caption{\textbf{Uniform, Symmetric and Hardware-Friendly Quantizers.}
    Illustration of the loss in quantization bins due to hardware friendly constraints.}
    \label{fig:quant_com}
\end{figure}

\paragraph{}
Figure \ref{fig:quant_com} illustrates uniform, symmetric and hardware-friendly 4-bit quantization grids for the same range of real numbers [-0.3,4.2] to be quantized. 
Specifically, the figure demonstrates how the symmetry and a power-of-two threshold constraints imply sub-optimal clipping ranges compared to the general uniform quantizer.
These clipping ranges lead to a loss in representation bins and thus increase the potential rounding noise.

%% file: files/bag_of_tricks.tex
\section{Method}\label{sec:method}
Given a trained floating point network and a representative dataset $D$ of independent and identically distributed samples, our aim is to quantize the network in post-training with hardware-friendly quantizers, namely that are uniform, symmetric and with power-of-two thresholds. 
Hardware Friendly Post Training Quantization (HPTQ) is a three-tier method for addressing this goal.
HPTQ consists of a pre-processing stage followed by activation quantization and weight quantization (see Fig. \ref{fig:cptq_flow}). 
In the resulting network, activations are quantized per tensor and weights are quantized per channel.

% For simplicity, in the rest of this section, when we refer to a quantizer $Q$ we assume it is signed, but it applies to both signed and unsigned quantizers.
%we quantized every convolution and fully connected layer and .
% Let $Q$ be a quantization function eme is parametrized by a pair (t, b) of threshold t and bit-width
% b. During training,

%Constrained Post Training Quantization (CPTQ) is a hardware friendly post training quantization method that consists of three stages: network preparation, activation quantization and weight quantization.   
%Our method assumes being given a trained floating point network and a representative dataset $D$ of independent and identically distributed samples. Figure \ref{fig:cptq_flow} details the overall flow of our method.
\input{files/flow}

%The CPTQ framework employs a bag of tricks for constraint post training quantization. The framework consists of a pair of new methods as well as original modifications to existing methods for improving the performance of a network under constrained quantization. 

%We analyzed the theoretical benefit of a constraint quantizer on the computational cost over different convolution network architectures.
%In addition, we analyze the computational overhead of unconstrained quantization compared with constrained one on key layers and evaluate the total overhead associated with several network architectures for different parameters such as width, resolution and depth.

\subsection{Pre-Processing}
The pre-processing stage consists of folding batch normalization layers into their preceding convolution layers \cite{jacob2018quantization}, collecting activation statistics using the representative dataset and finally removing outliers from the collected statistics. 

\paragraph{Batch-Normalization Folding.} 
A common technique to reduce model size and computational complexity is batch-normalization folding \cite{jacob2018quantization} (also known as batch-normalization fusing) in which batch-normalization layers are folded into the weights of their preceding convolution layers. 

\paragraph{Statistics Collection.}
In this stage we infer all of the samples in the representative dataset $D$ and collect activation statistics of each layer. Specifically, for each layer $l$ denote the collection of its activations over $D$ by $F_l\brackets{D}$.
Based on $F_l\brackets{D}$ we collect histograms for each tensor as well as the minimum, maximum and mean values per channel. In the reset of this work we assume that activation tensors $\matsym{X}\in\mathbf{R}^{h\times w \times c}$  have three dimensions where $h$, $w$ and $c$ are the height, weight and number of channels, respectively.

\begin{figure}[H]
     \centering
     \includegraphics[scale=0.3]{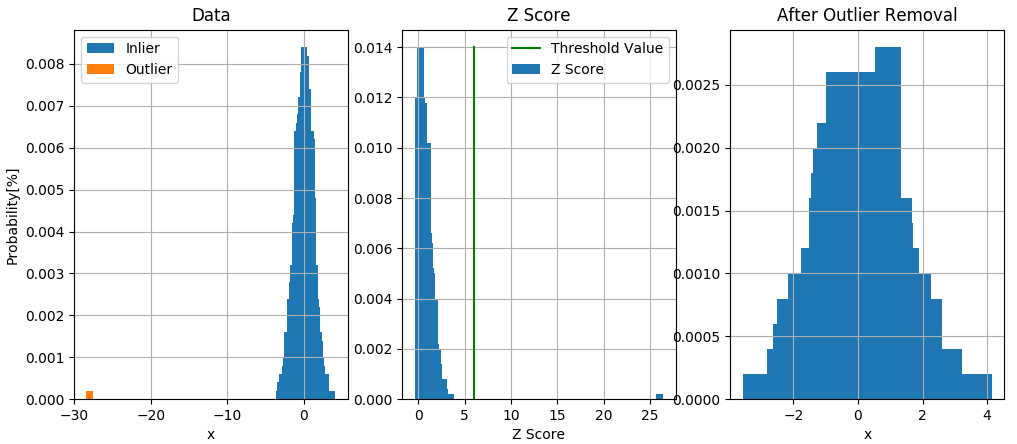}
     \caption{Outlier Removal. Left: an input data distribution. Middle: the respective distribution of absolute z-score values. Right: data distribution after outlier removal.}
     \label{fig:z_score}
\end{figure}

\paragraph{Outlier Removal.}
% TODO: In the current implmentation outlier removal is only applied for the threshold selection stage. Consider applying it in the pre-porcessing stage
In this step we filter out outliers in the activation histograms using the z-score approach described in \cite{aggarwal2015outlier}. 
Specifically, we remove histogram bins for which the absolute z-score value is larger than a predefined threshold. This implies that we restrict the range of each histogram bin to a predefined number of standard deviations from its activation mean value. 
See Figure \ref{fig:z_score} for an example.
%restriction to the standard deviation of the bin range. % from the activation mean value.
% We apply a known outlier removal method \cite{aggarwal2015outlier} that uses z-scores.
% Note that using this approach, outliers are removed from the histograms only if they satisfy the z-score condition. %opposed to common statistical methods in which values are always removed.
% To the best of our knowledge, we are the first to apply an outlier removal method for improving a post training quantization. 
Note that since this step updates the histograms, it applies only to the Threshold Selection step (see Figure \ref{fig:cptq_flow}).

% We have observed that in some cases the activation tensors have outliers in the collected statistics, which can affect the threshold selection results. Therefore, we apply this method before the threshold optimization as follows. 
% We calculate the z-score $z$ for each element:
% Let $h:B \rightarrow \mathbb{R}$ be the historgram of ... collected in.. we define the z-score vector $\overline{z}$ to be
% \begin{equation}\label{eq:z_score}
%   z=\frac{\abs{B-\mu}}{\sigma}
% \end{equation}
% where ... of the historgram
% where $\vectorsym{b}_{\matsym{X}}$ is the histogram bins of activation tensor $\matsym{X}$, $\mu$ is its empirical mean of  $\matsym{X}$ and $\sigma_{\matsym{x}}$ is its empirical standard deviation. Then, we remove the elements for which the z-score is above a predefined threshold $z_{th}$.

\subsection{Activation Quantization}
This stage consists of three steps: threshold selection, shift negative correction (SNC) and activation equalization.  
In the threshold selection step, we set power-of-two thresholds per tensor. The SNC step is a trick that improves the quantization of signed activation functions with a small negative range \cite{bhalgat2020lsq+}. 
In the activation equalization step we equalize the expected dynamic ranges of activation channels by applying a modified version of a technique that appears in \cite{nagel2019data}. 
% This technique is applied in our work to activations instead of weights and here, it is specific to symmetric and power-of-two quantization. 

% Note that the reason we quantize activations before weights is that the methods we use in the quantization of activations impose changes on weight values (e.g, Activation Equalization).

\paragraph{Threshold Selection.}   

%Note that in case of activation quantization, the search is performed on the tensor histograms, which can be collected using a representative dataset. \\

Given a fixed bit width $n_b$, our aim is to find a power-of-two threshold $t$ that minimizes the noise caused by the quantization of each layer $l$ in the network. Formally, for each layer $l$ in the network, our objective is to find a threshold $t$ that minimizes 
\begin{equation}\label{equ:err}
    ERR\brackets{t}=\frac{1}{n_s}\sum_{X\in F_l(D)}d\brackets{Q(\matsym{X},t,n_b),\matsym{X}},
\end{equation}
where $n_s$ is the size of the representative dataset, $F_l(D)$ is the collection of activation tensors in the $l$-th layer and $d$ is some error measurement.
% Due to the dependency of the activation values on the network inputs, which are unavailable in  post-training quantization, 

%Note that the error cannot be computed directly on the activation values due to their dependency on the network inputs. Instead, the error is estimated based on the collected statistic over the representative dataset $D$.

In an ablation study we examine the effect of several possible quantization error measurements on the actual task accuracy, including $\mathrm{L}_p$ Norms \cite{nahshan2019loss} and Kullback–Leibler (KL) divergence \cite{szymon}.
Our results show that Mean Square Error (MSE) \cite{nahshan2019loss} achieves the best performance (see Table \ref{tab:activation_t}). 
Thus, the objective of the threshold selection is to minimize
\begin{equation}
    ERR\brackets{t}=\frac{1}{n_s}\sum_{X\in F_l(D)}\brackets{Q(\matsym{X},t,n_b)-\matsym{X}}^2.
\end{equation}

% \begin{equation}
% \begin{split}
%     ERR\brackets{t}=\expectation{\mathrm{MSE}\brackets{Q(\matsym{X},t,n_b),\matsym{X}}}{\matsym{X}\sim D}&=\expectation{\brackets{Q(\matsym{X},t,n_b)-\matsym{X}}^2}{\matsym{X}\sim D}
% \end{split}
% \end{equation}

%Controlling the trade-off between two quantization noises, rounding noise \textit{vs.} clipping noise, is a key problem in quantization.

% A quantization error arises due to clipping and rounding operations in thequantization process.  Our threshold search tries to balance the impacts of these two factors.  It is performed in an iterative manner, leveraging the restriction onthe threshold to a discrete range of power-of-two values. 

% Note that quantization error arises due to clipping and rounding operations in the quantization process.  
In practice, we approximate a solution to this minimization problem by estimating the noise based on the histogram corresponding to layer $l$ collected in the Statistics Collection step above. % over the representative dataset $D$.
The restriction of the threshold to power-of-two values implies that the search space is discrete. 
Let  $M=\max\limits_{\matsym{X}\in F_l(D)}\max\limits_{i,j,k}\abs{\matsym{X}}_{i,j,k}$ be the maximal absolute value of an activation in $\matsym{X}$ over the representative dataset $D$ that was collected in the Statistics Collection step above and define the no-clipping threshold:
% $ that is, that is greater than the maximal absolute value of an activation in $X$ over the representative dataset $D$ that was collected in the Statistics Collection step above:
\begin{equation}\label{eq:no_clipping_a}
    t_{nc}=2^{\ceil{\log_2 M}}.
\end{equation}  
Note that the clipping noise induced by the threshold $t_{nc}$ is zero and that for any power-of-two threshold larger than $t_{nc}$, the noise is increased.
Thresholds smaller than $t_{nc}$ may reduce the noise, albeit, at the cost of increasing the clipping noise. 
% Empirically, this trade-off is preferable up to some limit \cite{NEURIPS2019_c0a62e13,choukroun2019low}. 
% \begin{equation}\label{eq:no_clipping_a}
%     t_{nc}=\max\limits_{\matsym{X}\in F_l(D)}2^{\ceil{\log_2\brackets{\max \abs{\matsym{X}}}}}
% \end{equation}  
% where, we denote $\abs{\cdot}$ as an element-wise absolute function throughout this work. 
%Similarly to \cite{jacob2018quantization}, we begin with a no-clipping threshold that satisfies both the symmetry and power-of-two constraints. Specifically, given a tensor, symmetry is achieved by taking its absolute value. Second, the power-of-two constraint is satisfied by ceiling the maximal value to the nearest power-of-two value. This results in the following equation for a no-clipping threshold $t_{nc}$: 
Therefore, we search for a threshold minimizing the quantization error starting with $t_{nc}$ and iteratively decreasing it (see. Algorithm \ref{alg:threshold_selection}).

\begin{algorithm}[H]\label{alg:threshold_selection}
 \KwData{quantization error estimator ERR ; no-clipping threshold $t_{nc}$; bit-width $n_b$; $n$ iterations}
 \KwResult{$t$ threshold value }
 $e_{min}=\infty$ \;
 $t=t_{nc}$  \;
 \For{i in 0 to n}{
  $t_i=\frac{t_{nc}}{2^{i}}$ \;
  $e_i=ERR\brackets{t_i,n_b}$ \;
  \If{$e_i < e_{min}$}{
    $t=t_i$ \; 
    $e_{min}=e_i$
  }
 }
 return $t$
 \caption{Constraint threshold selection}
\end{algorithm}

\paragraph{Shift Negative Correction (SNC).}  
Recent works have shown benefits in using signed, non-linear activation functions, such as Swish \cite{ramachandran2017searching}, PReLU and HSwish \cite{howard2019searching}. However, a signed symmetric quantization of these functions can be inefficient due to differences between their negative and positive dynamic ranges.
The main idea in SNC is to reduce the quantization noise of an unsigned activation function with a small negative range (relatively to its positive range). 
This is done by adding a positive constant to the activation values (shifting its values) and using an unsigned quantizer with the same threshold. This effectively doubles the quantization grid resolution.
Note that shifting the values can imply added clipping noise on the one hand but reduced rounding noise on the other.

This step can be viewed as an adaptation to PTQ of a technique that appears in \cite{bhalgat2020lsq+}, where activations are shifted and scaled in order to match a given dynamic range of a quantizer. Here, we do not add scaling due to its implied added complexity.
Specifically, let $\phi$ be the activation function in some layer $l$ in the network, let $t$ be its threshold, calculated in the Threshold Selection step above and let
% $\rm{s}=\min\limits_{\matsym{X}\in F_l(D)}\min\limits_{i,j,k}\matsym{X}_{ijk}$ 
$\rm{s}=\min\limits_{\matsym{X}\in F_l(D)}\min\limits_{i,j,k}\matsym{X}_{i,j,k}$ 
be its minimal (negative) activation value over the representative dataset $D$, collected in the Statistics Collection step above.
If $\frac{|\rm{s}|}{t} < \alpha$ for a hyperparameter $\alpha$, then we replace $\phi$ with a shifted version $\tilde\phi = \phi + |s|$ and replace the signed quantizer with an unsigned quantizer followed by another shift operation as follows:
\begin{equation}
    Q^{\rm{s}}(\phi(\matsym{X}), t,n_b)\xrightarrow{}Q^{\rm{us}}(\tilde{\phi}(\matsym{X}), t,n_b)-|s|,
\end{equation}
where $Q^{\rm{s}}(\phi(\matsym{X}), t,n_b)$ is the signed quantizer, $Q^{\rm{us}}(\tilde{\phi}(\matsym{X}), t,n_b)$ is the unsigned quantizer and $n_b$ is the bit-width.
%and replace the signed quantizer $Q^{\rm{s}}(\phi(\matsym{X}), t,n_b)$ with an unsigned quantizer followed by another shift operation $Q^{\rm{us}}(\tilde{\phi}(\matsym{X}), t,n_b)+s$ where $n_b$ is the bit-width.
% and the addition of the scalar $\rm{s}$ to a tensor is done per element.
% \begin{equation}
% Q^{\rm{s}}(\tilde\phi(\matsym{X}),t,n_b) = Q^{\rm{us}}(\tilde\phi(\matsym{X}), t,n_b)+s 
% \end{equation}
% \begin{equation}
% Q^{\rm{s}}(\phi(\matsym{X}), t,n_b)=Q^{\rm{us}}(\tilde{\phi}(\matsym{X}), t,n_b)+s 
% \end{equation}
% where $Q^{\rm{s}}$ and $Q^{\rm{us}}$ are signed and unsigned quantizers and $n_b$ is the bit-width.
% The subtraction and addition of $\rm{s}$ to a tensor are done element-wise.
In practice, the last subtraction of $|\rm{s}|$ is folded into the following operation in the network.
% and in case of a convolution with padding, this requires a non-zero padding value.

\begin{figure}[H]
    \centering
    \includegraphics[scale=0.22]{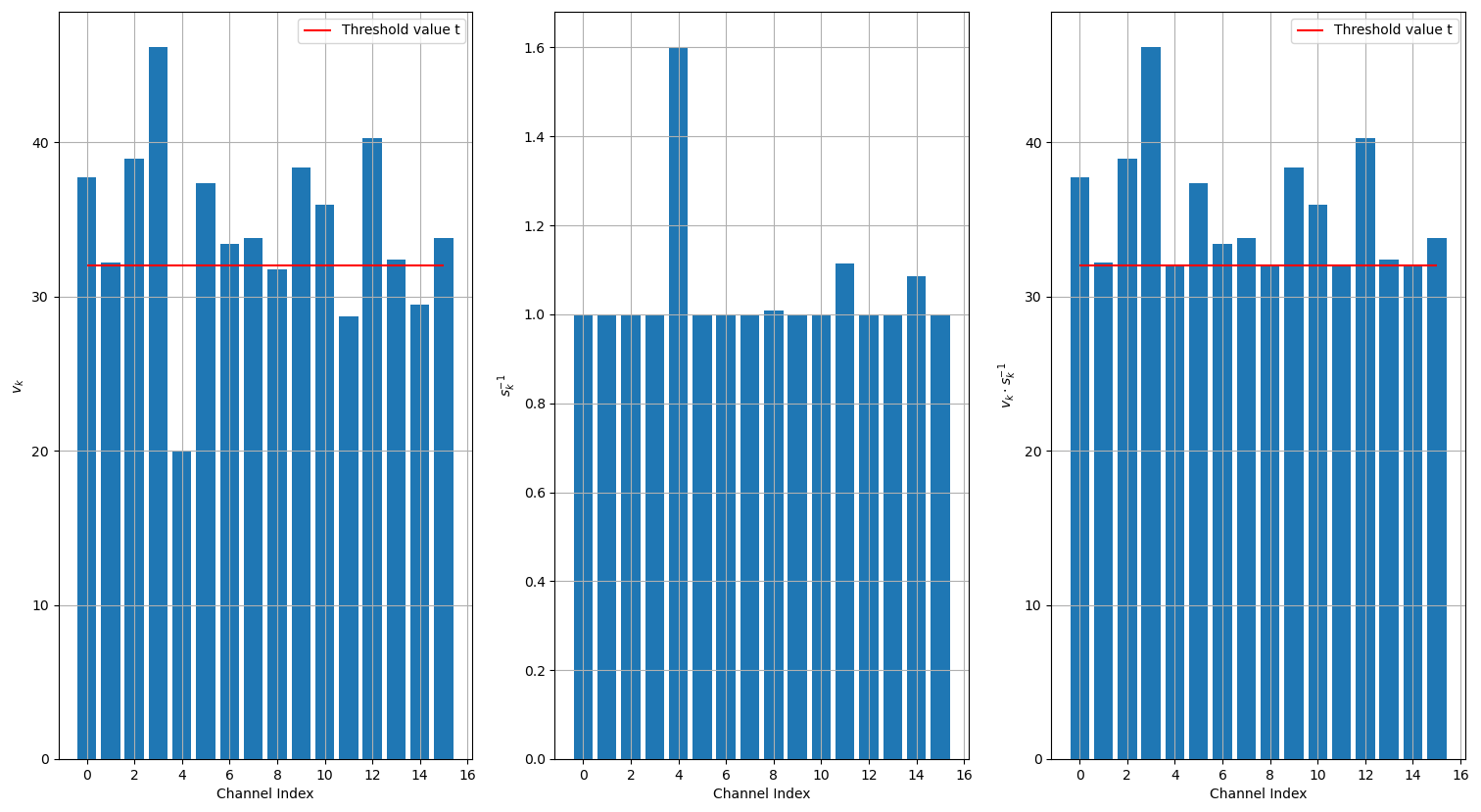}
    \caption{
    An example of Max Channel Equalization using \mbvtwo. 
    Left: the max value $\vectorsym{v}_{max}$ of each channel.
    Middle: the inverse scale factor $\frac{1}{\vectorsym{s}_k}$ for each channel $k$.
    Right: the max value of each channel after equalization using this scaling factor.
    }
    \label{fig:channels_corr}
\end{figure}

\paragraph{Activation Equalization.} 
% This approach has been suggested by \cite{nagel2019data,meller2019same} to improve weight quantization.
% In this step we equalize activation ranges based on weight equalization method that is presented in \cite{nagel2019data,meller2019same}. 
In this step, we equalize activation ranges per channel similarly to the methods presented in \cite{nagel2019data,meller2019same}. 
Here, we set the scale-per-channel factor according to the value of the threshold that is selected per-tensor. The motivation to use this scaling factor in order to equalize the activation ranges is to use the maximum range of the quantization bins for each channel (see Figure \ref{fig:channels_corr}).

%Relation between threshold and scaling -> threshold optimized best noise and then scale to match.

% The authors in \cite{nagel2019data} suggested to equalize the weight ranges by exploiting the scale equivariance property of activation functions. 
The authors in \cite{nagel2019data, meller2019same} suggest to perform channel equalization by exploiting the positive scale equivariance property of activation functions. 
% Here we use the same property to equalize the activation ranges instead of the weight ranges. 
It holds for any piece-wise linear activation function in its relaxed form:
%$\phi\brackets{\vectorsym{s}\odot\vectorsym{x}}=\vectorsym{s}\odot\hat{\phi}\brackets{\vectorsym{x}}$
${\phi}\brackets{\vectorsym{S}\vectorsym{x}}=\vectorsym{S}\hat{\phi}\brackets{\vectorsym{x}}$
where $\phi$ is a piece-wise linear function, $\hat{\phi}$ is its modified version that fits this requirement and $\vectorsym{S}=diag\brackets{\vectorsym{s}}$ is a diagonal matrix with $\vectorsym{s}_{k}$ denoting the scale factor for channel $k$.

The positive scaling equivariance can be applied on the following set of consecutive layers: a linear operation, a piece-wise linear function $\phi$ and an additional linear operation. This is demonstrated in the following equation:
\begin{equation}\label{eq:equalization}
\begin{split}
    \matsym{y} &=\matsym{W}_2\phi\brackets{\matsym{W}_1\matsym{x}+\vectorsym{b}_1}+\vectorsym{b}_2=\matsym{W}_2\phi\brackets{{\vectorsym{S}}{\vectorsym{S}^{-1}}\brackets{\matsym{W}_1 \matsym{x}+\vectorsym{b}_1}}+\vectorsym{b}_2\\
    &= \matsym{W}_2\vectorsym{S}\hat{\phi}({\vectorsym{S}^{-1}}\brackets{\matsym{W}_1 \matsym{x}+\vectorsym{b}_1})+\vectorsym{b}_2,
\end{split}
\end{equation}
where $\matsym{W}_1$ and $\vectorsym{b}_1$ are the first layer's weights and bias, $\matsym{W}_2$ and $\vectorsym{b}_2$ are the second layer's  weights and bias. Although Eq. \ref{eq:equalization}  demonstrates the case of fully-connected layers, it can be also extended for CNNs where the scaling is performed per channel.

% \begin{equation}
% \begin{split}
%     \matsym{Y} &=\matsym{W}_2\phi\brackets{\matsym{W}_1\matsym{X}+\vectorsym{b}_1}+\vectorsym{b}_2=\matsym{W}_2\phi\brackets{\frac{\vectorsym{s}}{\vectorsym{s}}\odot\brackets{\matsym{W}_1 \matsym{X}+\vectorsym{b}_1}}+\vectorsym{b}_2\\
%     &=\vectorsym{s}\odot  \matsym{W}_2\hat{\phi}(\frac{1}{\vectorsym{s}}\odot\brackets{\matsym{W}_1 \matsym{X}+\vectorsym{b}_1})+\vectorsym{b}_2
% \end{split}
% \end{equation}
% where $\vectorsym{s}$ is a scale vector, $\matsym{W}_1$ and $\vectorsym{b}_1$ are the first layer's kernel weights and bias, $\matsym{W}_2$ and $\vectorsym{b}_2$ are the second layer's kernel weights and bias.
% $\phi$ is a piece-wise non-linear function and $\hat{\phi}$ is its modified version to relax the positive scale equivariance requirement so that $\phi\brackets{\vectorsym{s}\odot\matsym{X}}=\vectorsym{s}\odot\hat{\phi}\brackets{\matsym{X}}$. 
% Here we show how channel equalization \cite{nagel2019data} can be used to overcome the performance degradation due to activation quantization. 
% We present two use cases of channel equalization: \textit{Max Channel Equalization} (to any quantization scheme) and \textit{ReLU6 Equalization} (specific to constraint quantization). 
We present a use case of channel equalization named \textit{Max Channel Equalization} which can be applied in any quantization scheme.
% we set the scale-per-channel factor according to the value of the threshold that is selected per-tensor
We assume that $\hat{\phi}$ is one of the following non-linear functions: ReLU, ReLU8 or PReLU. Given the quantization threshold $t$ of a non-linear function as well as the maximal activation value of the $k^{th}$ channel  $v_k=\max\limits_{\matsym{X}\in F_l(D)}\max\limits_{i,j}|\matsym{X}_{i,j,k}|$, where $\matsym{X}$ is the activation tensor of the $l^{th}$ layer, we set:
\begin{equation}
    \vectorsym{s}_k=\min\brackets{\frac{v_k}{t},1},
\end{equation}
so that the maximal value of each channel in tensor $\matsym{X}$ will be the threshold value (see Figure \ref{fig:channels_corr}).
% , since %$\hat{\phi}\brackets{\frac{1}{\vectorsym{s}}\odot\matsym{X}}=\frac{1}{\vectorsym{s}}\odot{\phi}\brackets{\matsym{X}}$.
%$\hat{\phi}\brackets{\vectorsym{S}^{-1}\vectorsym{x}}=\vectorsym{S}^{-1}{\phi}\brackets{\vectorsym{x}}$.

% In case of \textit{ReLU6 Equalization}, a threshold value of $8$ is typically used, since it is the smallest value that meets the power of two constraint. This results in a set of unused quantization bins (between 6 and 8), which might cause a precision degradation. To alleviate this and use all quantization bins, we set $\vectorsym{s}_k=\frac{6}{8}$ and  $\hat{\phi}\brackets{\matsym{X}}=\mathrm{ReLU8}\brackets{\matsym{X}}$.\\

\subsection{Weight Quantization}
In the Weight Quantization stage we quantize the network's weights.
It was shown in \cite{krishnamoorthi2018quantizing,rastegari2016xnor} that weight quantization with scaling per channel improves accuracy.
Moreover, this work presents an efficient dot product and convolution implementation supporting per-channel quantization. 
Our Weight Quantization stage consists of per-channel threshold selection and bias correction \cite{nagel2019data}.

%A major challenge in weight quantization is dealing with weight distribution, which varies dramatically across channels. The authors of \cite{krishnamoorthi2018quantizing,rastegari2016xnor} suggested to calculate a scaling factor for each channel and showed an improvement in accuracy. Moreover, they showed that scaling per channel quantization has an efficient dot product and convolution implementation. 

\paragraph{Threshold Selection.} 
As noted above, weight quantization is performed per-channel. Its thresholds are selected similarly to activation thresholds (see Algorithm \ref{alg:threshold_selection}). However, a key difference is that here the search is performed directly on the weight values, opposed to the statistical values that are used for activation.
More precisely, given the weights $\vectorsym{w}\in\mathbb{R}^{n}$ of some channel in the network, the initial no-clipping threshold is 
\begin{equation}\label{eq:no_clipping_w}
    t_{nc}=2^{\ceil{\log_2\brackets{\max\limits_i \abs{\vectorsym{w}_i}}}},
\end{equation}
where $\vectorsym{w}_i\in\mathbb{R}$ are the entries of $\vectorsym{w}$. 
Additionally, the error induced by a threshold $t$ is
\begin{equation}
    ERR\brackets{t}=MSE(Q(\vectorsym{w},t,n_b),\vectorsym{w})=\frac{1}{n}\sum_i \brackets{Q(\vectorsym{w}_i,t,n_b)-\vectorsym{w}_i}^2.
\end{equation} 
Note that as with activations, MSE is selected as an error measurement since it yields the best performance (see Table \ref{tab:weights_t}). 

\paragraph{Bias Correction.} Quantization of weights induce bias shifts to activation means that may lead to detrimental behaviour in the following layers \cite{nagel2019data,finkelstein2019fighting}.  
Explicitly, let $\vectorsym{y}=\matsym{W}\vectorsym{x}+\vectorsym{b}$ be the floating point output of a fully connected layer where $\vectorsym{x}, \matsym{W}, \vectorsym{b}$ are the floating-point input activation, weight and bias, respectively. 
Denote the quantized weights of the layer by $\tilde{\matsym{W}} = Q(\matsym{W},t,n_b)$
and the corresponding output by
$\tilde{\vectorsym{y}}=\tilde{\matsym{W}}\vectorsym{x}+\vectorsym{b}$.
% The quantized output of the same fully connected layer is given by $\tilde{\vectorsym{y}}=Q(\matsym{W},t,n_b)\vectorsym{x}+\vectorsym{b}=\tilde{\matsym{W}}\vectorsym{x}+\vectorsym{b}$ where  $\tilde{\matsym{W}}$ is the quantized  weight tensors with threshold $t$ and $n_b$ bits.  
The induced bias shift $\expectation{\vectorsym{y}}{}-\expectation{\tilde{\vectorsym{y}}}{}$ can be expressed as follows:
\begin{equation}\label{eq:bias_correction}
    \expectation{\vectorsym{y}}{}-\expectation{\tilde{\vectorsym{y}}}{}=\expectation{\matsym{W}-\tilde{\matsym{W}}}{}\expectation{\vectorsym{x}}{}=\epsilon\expectation{\vectorsym{x}}{}.
\end{equation}
% Note that 
% $\epsilon = \expectation{\matsym{W}-\tilde{\matsym{W}}}{}$ is the weight quantization error. 
Several works propose approaches to correct the quantization induced bias. 
These include using batch-normalization statistics \cite{nagel2019data}, micro training  \cite{finkelstein2019fighting} and applying scale and shift per channel \cite{NEURIPS2019_c0a62e13}.

We adopt the solution in \cite{nagel2019data}, in which the bias shift is fixed by modifying the layer's bias vector
\begin{equation}
\tilde{\vectorsym{b}}=\vectorsym{b}-\epsilon\expectation{\vectorsym{x}}{},
\end{equation}
where $\expectation{\vectorsym{x}}{}$ is the per channel empirical mean obtain in the Statistic Collection stage above. 
% That is, the $k$-th entry in $\expectation{\vectorsym{x}}{}$ is equal to
% $\frac{1}{n_s}\sum_{\vectorsym{x}\in F_l(D)}\vectorsym{x}_k$. 
Note that although the above is written for a fully connected layer, it applies to convolutional layers as well, as shown in \cite{nagel2019data}.
% , where the mean is performed per channel.

%This method was introduced by \cite{finkelstein2019fighting,nagel2019data} and can be applied to any weight quantization scheme. The authors showed that the mean of a layer’s output suffers from a bias due to weight quantization. This bias can be expressed by:
%\begin{equation}
%    \expectation{\matsym{Y}}{}-\expectation{\tilde{\matsym{Y}}}{}=\expectation{\matsym{W}-\tilde{\matsym{W}}}{}\expectation{\matsym{X}}{}=\epsilon \cdot \expectation{\matsym{X}}{}
%\end{equation}
%where $\matsym{X}$ is the input activation tensor, $\matsym{Y}$ and $\matsym{W}$ are the floating-point activation and weight tensors, $\tilde{\matsym{Y}}$ and $\tilde{\matsym{W}}$ are the quantized activation and weight tensors and $\epsilon=\matsym{W}-\tilde{\matsym{W}}$ is the weight quantization error. In the literature several approaches to correct the induced bias are proposed. For example, by batch-normalization statistics \cite{nagel2019data}, micro training  \cite{finkelstein2019fighting} or by applying scale and shift per channel \cite{NEURIPS2019_c0a62e13}.

%% file: files/flow.tex
\tikzset{doc/.style={document,fill=blue!10,draw,thin,minimum
height=1.2cm,align=center},
pics/.cd,
pack/.style={code={%
\draw[fill=blue!50,opacity=0.2] (0,0) -- (0.5,-0.25) -- (0.5,0.25) -- (0,0.5) -- cycle;
\draw[fill=blue!50,opacity=0.2] (0,0) -- (-0.5,-0.25) -- (-0.5,0.25) -- (0,0.5) -- cycle;
\draw[fill=blue!60,opacity=0.2] (0,0) -- (-0.5,-0.25) -- (0,-0.5) -- (0.5,-0.25) -- cycle;
\draw[fill=blue!60] (0,0) -- (0.25,0.125) -- (0,0.25) -- (-0.25,0.125) -- cycle;
\draw[fill=blue!50] (0,0) -- (0.25,0.125) -- (0.25,-0.125) -- (0,-0.25) -- cycle;
\draw[fill=blue!50] (0,0) -- (-0.25,0.125) -- (-0.25,-0.125) -- (0,-0.25) -- cycle;
\draw[fill=blue!50,opacity=0.2] (0,-0.5) -- (0.5,-0.25) -- (0.5,0.25) -- (0,0) -- cycle;
 \draw[fill=blue!50,opacity=0.2] (0,-0.5) -- (-0.5,-0.25) -- (-0.5,0.25) -- (0,0) -- cycle;
\draw[fill=blue!60,opacity=0.2] (0,0.5) -- (-0.5,0.25) -- (0,0) -- (0.5,0.25) -- cycle;
}}}
\begin{figure}[H]
    \centering
    \begin{tikzpicture}[font=\small,every label/.append style={font=\small,align=center}]
        %\node[cylinder, cylinder uses custom fill, cylinder end fill=blue!25, cylinder body fill=blue!40,shape border rotate=90, aspect=0.4,minimum width=1cm,minimum height=1.4cm, text width=1.4cm,align=center,font=\tiny](net){Input Network};
        \node (net) [startstop,font=\tiny,text width=1.4cm] {Input Network};
        \node[below=0.6cm of net,block,fill=green!15, text width=1.4cm,font=\tiny,align=center] (bf) {Batch\\Normalization Folding};

        \node[below=0.6cm of bf,block,fill=green!15, text width=1.4cm,font=\tiny,align=center] (at) {Threshold Selection};
       
        \node[below=0.3cm of at,block,fill=green!15, text width=1.4cm,font=\tiny,align=center] (snc) {Shift\\Negative Correction};
        \node[below=0.3cm of snc,block,fill=green!15, text width=1.4cm,font=\tiny,align=center] (ae) {Activation Equalization};
        %\node[fit=(Store) (Router)](fit1){};
        \node[left=0.5cm of bf,block,fill=green!15, text width=1.4cm,font=\tiny,align=center] (sc) {Statistics Collection};
        
         \node[left=0.5cm of sc,block,fill=green!15, text width=1.4cm,font=\tiny,align=center] (zs) {Outlier Removal};
        \node[below=0.5cm of zs,tape, draw,thin, tape bend top=none,fill=purple!20,double copy shadow,minimum height=1.5cm, text width=1.5cm,font=\tiny,align=center](sd) {Statistics Data};
        \node[above=0.6cm of sc,cylinder, cylinder uses custom fill, cylinder end fill=blue!25, cylinder body fill=blue!40,shape border rotate=90, aspect=0.4,minimum width=1cm,minimum height=1.4cm, text width=1.5cm,align=center,font=\tiny](ds) {Representative Dataset $D$};

        \node[below=0.6cm of ae,block,fill=green!15, text width=1.4cm,font=\tiny,align=center] (wt) {Threshold Selection};
        \node[below=0.3cm of wt,block,fill=green!15, text width=1.4cm,font=\tiny,align=center] (bc) {Bias\\Correction};
        %\node[below=0.3cm of bc,cylinder, cylinder uses custom fill, cylinder end fill=blue!25, cylinder body fill=blue!40,shape border rotate=90, aspect=0.4,minimum width=1cm,minimum height=1.4cm, text width=1.4cm,align=center,font=\tiny](qnet){Quantized Network};
        
        \node (qnet) [startstop,below=0.3cm of bc,font=\tiny,text width=1.4cm] {Quantized Network};
        \begin{pgfonlayer}{background}
            \node[draw,rounded corners,fit=(at) (ae) (snc),fill=black!20,inner sep=5pt,label={[label distance=-1.8cm,text depth=0.4cm,rotate=-90] right:{Activation Quantization}}](fit2){};
            \node[draw,rounded corners,fit=(wt) (bc),fill=yellow!30,inner sep=5pt,label={[label distance=-1.6cm,text depth=0.4cm,rotate=-90] right:{Weight Quantization}}](fit2){};
            \node[draw,rounded corners,fit=(bf) (zs) (sc),fill=red!30,inner sep=5pt,label={[label distance=-1.2cm,text depth=0.4cm,rotate=-90] right:{Pre-Processing}}](fit2){};
        \end{pgfonlayer}
        \node[dummy,below=0.825 of ae] (d) {};
        
        %\node[doc,right=1cm of fit2] (js) {app.js};
        %\node[doc,above right=1cm of js] (Server) {Server\\ entity};
        %\node[doc,below right=1cm of js] (Client) {Client\\ entry};
        \draw[->,dotted,line width=0.25mm] (zs)  -- (sd);
        %\draw[to-to,dashed,line width=0.25mm,xshift=0.1 cm](sd) |- (snc);
        \draw[->,dashed,line width=0.25mm](sd) |- ([yshift=0.1cm] snc.west);
        \draw[-<,dashed,line width=0.25mm](sd) |- ([yshift=-0.1cm] snc.west);
        
        \draw[->,dashed,line width=0.25mm](sd) -- (at);
        \draw[->,dashed,line width=0.25mm](sd) |- ([yshift=0.1cm]ae.west);
        \draw[-<,dashed,line width=0.25mm](sd) |- ([yshift=-0.1cm]ae.west);
        
        \draw[->,dashed,line width=0.25mm](sd) |- (bc);
        %\draw[->,dashed,line width=0.25mm](sd) |- (bc);
        %\draw[dashed,line width=0.25mm](d) -- (bc);
        
        \draw[->,line width=0.25mm] (net) -- (bf);    
        %\draw[->,line width=0.25mm] (net) -- (bf);
        \draw[->,line width=0.25mm] (bf) -- (sc);
        \draw[->,line width=0.25mm] (bf) -- (at);
        \draw[->,line width=0.25mm] (at) -- (snc);
        \draw[->,line width=0.25mm] (snc) -- (ae);
        \draw[->,line width=0.25mm] (ae) -- (wt);
        \draw[->,line width=0.25mm] (wt) -- (bc);
        \draw[->,line width=0.25mm] (bc) -- (qnet);
        
        \draw[->,dotted,line width=0.25mm] (sc) -- (zs);
        \draw[->,dotted,line width=0.25mm] (ds) -- (sc);
    \end{tikzpicture}
    \caption{The HPTQ framework. Dashed lines represent statistical information passing, which include also their updates, dotted lines represent data passing and solid lines represent an updated network. }
    \label{fig:cptq_flow}
\end{figure}

%% file: files/experimental_results.tex
\section{Experimental Results}\label{sec:experimental}
In this section we evaluate the performance of HPTQ with 8-bit quantization over different tasks and a variety of network architectures. 
The experiments are divided into two parts. The first part presents an overall performance comparison to the floating point baseline as well as to state-of-the-art quantization approaches. The second part presents an ablation study that analyzes the influence of each technique in HPTQ separately.

% an overall performance comparison to state-of-the-art methods and to floating point baselines and an ablation study that analyzes the influence of each technique in HPTQ separately.
% The results show that competitive results can be obtained under hardware friendly constraints of uniform, symmetric 8-bit quantization with a power-of-two threshold.

% We examine four different tasks: image classification, object detection, semantic segmentation and pose estimation. 
% A comprehensive performance comparison of HPTQ with other PTQ and QAT quantization methods is presented. 
% % We compare the performance of HPTQ with other PTQ and QAT quantization methods and show competitive results. 
% In addition, we perform an ablation study that analyzes the influence of each technique in HPTQ separately. 
% We evaluate the performance of HPTQ with 8-bit quantization for both activations and weights on classification and segmentation tasks. 

\subsection{Overall Performance Evaluation}
We evaluate the performance of HPTQ on four different tasks: image classification, object detection, semantic segmentation and pose estimation.
For each task, we present a comparison between the performance of models quantized by HPTQ and their floating point baselines. Furthermore, for classification and segmentation we provide a comprehensive performance comparison of HPTQ with both PTQ and QAT state-of-the art quantization methods.

% in comparison to state-of-the-art quantization methods.
% On these tasks, we compare HPTQ with different quantization methods including QAT and PTQ methods. 
We use the same set of hyper-parameters for all our experiments.
Specifically, the number of image samples in the representative dataset $D$ is 500. 
The z-score threshold in the outlier removal step is $z_{th}=24$.
The SNC threshold is $\alpha=0.25$. 
Last, for both activations and weights, the number of iterations performed in Algorithm \ref{alg:threshold_selection} in the threshold selection search is set to $n=10$. One should note that fine-tuning the hyper-parameters per network may lead to further improvement.
In all of the tables below $\Delta$ is the difference between the performance of the floating point model and the quantized model, \textit{PC} indicates the use of weights per channel quantization and \textit{PoT} indicates power-of-two thresholds.

%four aforementioned tasks with a wide variety of backbones. The results show competitive performance of HPTQ with a constraint quantizer. 

% \textit{G} indicates the use of gradients during the post training

\paragraph{Classification.} 
We evaluate HPTQ on the ImageNet classification task \cite{deng2009imagenet}  using \mbvone, \mbvtwo and \res architectures\footnote{\url{https://www.tensorflow.org/api_docs/python/tf/keras/applications}}. 
Tables \ref{table:mbv1_com}, \ref{table:mbv2_com} and \ref{table:res_com} present comparisons of HPTQ with other quantization methods, both PTQ and QAT, for the three architectures. The results show that HPTQ achieves competitive performance despite the hardware friendly constraints.
%In same case HPTQ achieve better accuracy than QAT methods , 
%Comparison HPTQ to QAT methods shows that the most advanced methods achieve better accuracy and even zero loss. 
%These are methods that learn the quantization thresholds  during supervised training. 
%However, methods that do not train their threshold value achieve lower accuracy than HPTQ.
In the tables below \textit{F-Acc} is the floating point accuracy and \textit{Q-Acc} is the accuracy of the quantized model.

\begin{table}[H]
\centering
\caption{ImageNet classification \cite{deng2009imagenet} with \mbvone  }
\label{table:mbv1_com}
\begin{tabular}{|c|l|c|c|l|l|l|}
\hline
\multicolumn{1}{|c|}{\textbf{Type}}                                           & \multicolumn{1}{|c|}{\textbf{Method}}          & \multicolumn{1}{|c|}{\textbf{PC}}              & \multicolumn{1}{|c|}{\textbf{PoT}}        & \multicolumn{1}{|c|}{\textbf{F-Acc}} & \multicolumn{1}{|c|}{\textbf{Q-Acc}} & \multicolumn{1}{|c|}{\textbf{$\Delta$}} \\ \hline
\multirow{2}{*}{\rotatebox[origin=c]{90}{QAT}} & \qt             & \xmark      & \xmark     & 70.9  & 70.0    & 0.9      \\ \cline{2-7} 
                                               & \tqt            & \xmark      & \checkmark & 71.1  & 71.1  & 0.0        \\ \hline
\multirow{5}{*}{\rotatebox[origin=c]{90}{PTQ}} & \ssbd           & \xmark      & \xmark     & 70.9  & 69.95 & 0.95     \\ \cline{2-7} 
                                               & \Krishnamoorthi & \checkmark  & \xmark     & 70.9  & 70.3  & 0.6      \\ \cline{2-7} 
                                               & \wu             & \checkmark  & \xmark     & 71.88 & 70.39 & 1.49     \\ \cline{2-7} 
                                               & \lee            & \xmark      & \xmark     & 69.5  & 68.84 & 0.66     \\ \cline{2-7} 
                                               & \textbf{\cptq}           & \checkmark  & \checkmark & \textbf{70.55} & \textbf{70.41} & \textbf{0.14}     \\ \hline
\end{tabular}
\end{table}

\begin{table}[H]
\centering
\caption{ImageNet classification \cite{deng2009imagenet} with \mbvtwo}
\label{table:mbv2_com}
\begin{tabular}{|c|l|c|c|l|l|l|}
\hline
\multicolumn{1}{|c|}{\textbf{Type}}                                           & \multicolumn{1}{|c|}{\textbf{Method}}          & \multicolumn{1}{|c|}{\textbf{PC}}              & \multicolumn{1}{|c|}{\textbf{PoT}}        & \multicolumn{1}{|c|}{\textbf{F-Acc}} & \multicolumn{1}{|c|}{\textbf{Q-Acc}} & \multicolumn{1}{|c|}{\textbf{$\Delta$}} \\ \hline
\multirow{3}{*}{\rotatebox[origin=c]{90}{QAT}}  & \qt                     & \xmark          & \xmark     & 71.9                   & 70.9   & 1.0        \\ \cline{2-7} 
                                                & \rvquant                & \xmark          & \xmark     & 70.10                 & 70.29 & -0.19    \\ \cline{2-7} 
                                                & \tqt                    & \xmark          & \checkmark & 71.7                   & 71.8   & -0.10     \\ \hline
\multirow{10}{*}{\rotatebox[origin=c]{90}{PTQ}} & \adaquant               & \xmark      & \xmark     & 73.03                  & 73.03  & 0.0        \\ \cline{2-7} 
                                                 & \zeroq                  & \xmark      & \xmark     & 73.03                  & 72.91  & 0.12     \\ \cline{2-7} 
                                                & \ssbd                   & \xmark        & \xmark     & 71.9                   & 71.29  & 0.61     \\ \cline{2-7} 
                                                & \wu                     & \checkmark     & \xmark     & 71.88                  & 71.14  & 0.74     \\ \cline{2-7} 
                                                & \Krishnamoorthi         & \checkmark      & \xmark     & 71.9                   & 69.7   & 2.2      \\ \cline{2-7} 
                                                & \multirow{2}{*}{\nagel} & \xmark          & \xmark     & \multirow{2}{*}{71.72} & 70.99  & 0.73     \\ \cline{3-4} \cline{6-7} 
                                                &                         & \checkmark      & \xmark     &                        & 71.16  & 0.56     \\ \cline{2-7} 
                                                & \dfq                    & \xmark          & \xmark     & 71.72                  & 70.92  & 0.8      \\ \cline{2-7} 
                                                & \lee                    & \xmark          & \xmark     & 71.23                  & 69.5   & 1.73     \\ \cline{2-7} 
                                                & \textbf{\cptq}                   & \checkmark      & \checkmark & \textbf{71.812}                 & \textbf{71.46}  & \textbf{0.352}    \\ \hline
\end{tabular}
\end{table}

\begin{table}[H]
\centering
\caption{ImageNet classification \cite{deng2009imagenet} with \res}
\label{table:res_com}
\begin{tabular}{|c|l|c|c|l|l|l|}
\hline
\multicolumn{1}{|c|}{\textbf{Type}}                                           & \multicolumn{1}{|c|}{\textbf{Method}}          & \multicolumn{1}{|c|}{\textbf{PC}}              & \multicolumn{1}{|c|}{\textbf{PoT}}        & \multicolumn{1}{|c|}{\textbf{F-Acc}} & \multicolumn{1}{|c|}{\textbf{Q-Acc}} & \multicolumn{1}{|c|}{\textbf{$\Delta$}} \\ \hline
\multirow{6}{*}{\rotatebox[origin=c]{90}{QAT}}  & \qt                              & \xmark         & \xmark     & 76.4                   & 74.9   & 1.5      \\ \cline{2-7} 
                                                & \rvquant                         & \xmark         & \xmark     & 75.92                  & 75.67  & 0.25     \\ \cline{2-7} 
                                                & \hawq                            & \checkmark    & \xmark     & 77.72                  & 77.58  & 0.14     \\ \cline{2-7} 
                                                & \lsq                             & \xmark          & \xmark     & 76.9                   & 76.8   & 0.1      \\ \cline{2-7} 
                                                & \tqt                             & \xmark          & \checkmark & 76.9                   & 76.5   & 0.4      \\ \cline{2-7} 
                                                & \faq                             & \xmark          & \xmark     & 75.4                   & 75.4   & 0.0        \\ \hline
 \multirow{10}{*}{\rotatebox[origin=c]{90}{PTQ}} & \zeroq                           & \xmark      & \xmark     & 77.72                  & 77.67  & 0.05     \\ \cline{2-7} 
                                                & \ocs                             & \xmark         & \xmark     & 76.1                   & 75.9   & 0.2      \\ \cline{2-7} 
                                                & \ssbd                            & \xmark          & \xmark     & 75.2                   & 74.95  & 0.25     \\ \cline{2-7} 
                                                & \he                              & \xmark          & \xmark     & 75.3                   & 75.03  & 0.27     \\ \cline{2-7} 
                                                & \wu                              & \checkmark     & \xmark     & 76.16                  & 76.05  & 0.11     \\ \cline{2-7} 
                                                & \multirow{2}{*}{\nagel}          & \xmark         & \xmark     & \multirow{2}{*}{76.07} & 75.87  & 0.2      \\ \cline{3-4} \cline{6-7} 
                                                &                                  & \checkmark      & \xmark     &                        & 75.88  & 0.19     \\ \cline{2-7} 
                                                & \multirow{2}{*}{\Krishnamoorthi} & \xmark         & \xmark     & \multirow{2}{*}{75.2}  & 75.00     & 0.20      \\ \cline{3-4} \cline{6-7} 
                                                &                                  & \checkmark     & \xmark     &                        & 75.1   & 0.1      \\ \cline{2-7} 
                                                & \textbf{\cptq}                            & \checkmark      & \checkmark & \textbf{75.106}                 & \textbf{75.018} & \textbf{0.088}    \\ \hline
\end{tabular}
\end{table}

\paragraph{Semantic Segmentation.}
We evaluate HPTQ on Pascal VOC \cite{everingham2010pascal} using DeepLab V3\footnote{\url{https://github.com/tensorflow/models/blob/master/research/deeplab/g3doc/model_zoo.md}} \cite{chen2017rethinking} with \mbvtwo as a backbone. Table \ref{tab:deeplab} shows that HPTQ achieves competitive results compared to other PTQ methods.

\begin{table}[H]
\caption{Semantic segmentation on Pascal VOC \cite{everingham2010pascal} using DeepLab V3 with \mbvtwo as a backbone. \textit{F-mIoU} is the floating point  mean Intersection-over-Union (mIoU) and \textit{Q-mIoU} is the mIoU of the quantized model.}
\label{tab:deeplab}
\centering
\begin{tabular}{|c|l|c|c|l|l|l|}
\hline
\multicolumn{1}{|c|}{\textbf{Type}}                                           &  \multicolumn{1}{|c|}{\textbf{Method}}                  &  \multicolumn{1}{|c|}{\textbf{PC}}             &  \multicolumn{1}{|c|}{\textbf{PoT}}        &  \multicolumn{1}{|c|}{\textbf{F-mIoU}}                  & \multicolumn{1}{|c|}{\textbf{Q-mIoU}} & \multicolumn{1}{|c|}{\textbf{$\Delta$}} \\ \hline
\multirow{4}{*}{\rotatebox[origin=c]{90}{PTQ}} & \dfq                    & \xmark      & \xmark     & 72.45                  & 72.33 & 0.12     \\ \cline{2-7} 
                                               & \multirow{2}{*}{\nagel} & \xmark     & \xmark     & \multirow{2}{*}{72.94} & 72.44 & 0.50      \\ \cline{3-4} \cline{6-7} 
                                               &                         & \checkmark  & \xmark     &                        & 72.27 & 0.67     \\ \cline{2-7} 
                                               & \textbf{\cptq}                   & \checkmark  & \checkmark & \textbf{75.57}                  & \textbf{75.38} & \textbf{0.19}     \\ \hline
\end{tabular}
\end{table}

% \subsection{Additional Results}
% Here we provide additional results of HPTQ on pose estimation and object detection task as well as more results on image classification task. 
%Here we evaluate the performance of the entire HPTQ framework with 8-bit quantization for both activations and weights on the four aforementioned tasks with a wide variety of backbones. The results show competitive performance of HPTQ with a constraint quantizer. 

%In all our experiments we use the MSE cost function for both activation and weight thresholds. Moreover, for weight quantization we utilize all methods shown in Section \ref{sec:method}: per channel threshold and bias correction. In addition, for activation quantization... All these selections are motivated by the ablation study below.

%Based on these observations, we utilize both the per channel threshold and bias correction, since per channel threshold improve the accuracy and bias correction can have a large improve to the performance or addition a small noise depending on the network. 

%The results show competitive performance of HPTQ  with a constraint quantizer. 

\paragraph{Object Detection.} 
We evaluate HPTQ on COCO  \cite{lin2014microsoft} using the SSD detector \cite{liu2016ssd} with several backbones\footnote{\url{https://github.com/tensorflow/models/blob/master/research/object_detection/g3doc/tf2_detection_zoo.md}}. 
HPTQ achieves similar Mean Average Precision (mAP) to the floating point baseline as demonstrated in Table  \ref{table:det}.
% Table \ref{table:det} shows the Mean Average Precision (mAP) achieved by HPTQ in comparison with the floating point baseline, which demonstrate similar performance.

\begin{table}[H]
\centering
\caption{Object detection results with HPTQ on COCO  \cite{lin2014microsoft} using \mbvtwo and \res as backbones. \textit{F-mAP} is the floating point mAP  and \textit{Q-mAP} is the mAP of the quantized model.}
\label{table:det}
\begin{tabular}{|l|l|l|}
\hline
\multicolumn{1}{|c|}{\textbf{Model}}                           & \multicolumn{1}{|c|}{\textbf{F-mAP}} & \multicolumn{1}{|c|}{\textbf{Q-mAP}} \\ \hline
SSD \mbvtwo 320x320  & 20.2  & 20.21     \\ \hline
SSD  \mbvtwo FPN Lite 320x320        & 22.2  & 21.93     \\ \hline
SSD \res V1 FPN 640x640     & 34.3  & 34.3      \\ \hline
\end{tabular}
\end{table}

%\paragraph{Semantic Segmentation.}
%We evaluate HPTQ on the semantic segmentation task using DeepLab V3\footnote{\url{https://github.com/tensorflow/models/blob/master/research/deeplab/g3doc/model_zoo.md}} \cite{chen2017rethinking} that has been trained on Pascal VOC \cite{everingham2010pascal}. Table \ref{table:seg} shows the Mean Intersection over Union (mIoU) achieved by HPTQ in comparison with the floating point baseline, which demonstrate competitive performance.

%\begin{table}[H]
%\centering
%\caption{HPTQ Results for the semantic segmentation task, using DeepLab V3 with two different backbones: \mbvtwo and \xecption}
%\label{table:seg}
%\begin{tabular}{|c|c|c|c|}
%\hline
%Network                     & Backbone     & Float & Quantized \\ \hline
%\multirow{2}{*}{DeepLab V3} & \mbvtwo &75.57 & 75.38     \\ \cline{2-4} 
%                            & \xecption     & 82.20 & 81.86     \\ \hline
%\end{tabular}
%\end{table}

\paragraph{Pose-Estimation.} 
We evaluate HPTQ on the single-person pose estimation task using LPN network \cite{zhang2019simple} on the LIP (Look into Person) dataset \cite{liang2018look}. 
We use the PCKh metric \cite{liang2018look} for evaluation, which is the head-normalized probability of correct keypoints. 
HPTQ achieves similar performance to the floating point baseline with only a slight degradation from 81.65 to 81.53 PCKh.

% \begin{table}[H]
% \centering
% \caption{pCKH achieved by HPTQ for the single-person pose estimation task.}
% \label{table:pose}
% \begin{tabular}{|c|c|c|}
% \hline
% Network & F-Acc & Q-Acc \\ \hline
% LPN     & 81.65 & 81.53     \\ \hline
% \end{tabular}
% \end{table}

\subsection{Ablation Study}\label{sec:ablation}

We provide an ablation study of HPTQ's performance on the ImageNet classification task \cite{deng2009imagenet} using eleven networks\footnote{\url{https://www.tensorflow.org/api_docs/python/tf/keras/applications}}. 
The study is divided into two parts analyzing activation quantization and weight quantization. 

Table \ref{table:atype_effect} compares the performance of HPTQ between four cases: full floating-point, activation quantization, weight quantization and joint quantization of both. 
The comparison shows that activation quantization causes a larger degradation in performance compared to weight quantization, especially for EfficientNet with Swish activations functions. 
This might be due to the fact that activation equalization is not applied for these activations.
% especially for networks with Swish activations (EfficientNet). 

\begin{table}[H]
\centering
\caption{ImageNet classification \cite{deng2009imagenet} accuracy with HPTQ in four cases: full floating-point, activation quantization, weight quantization and both activation and weight quantization.}
\label{table:atype_effect}
\begin{tabular}{|l|l|l|l|l|}
\hline
\multicolumn{1}{|c|}{\textbf{Network}} & \multicolumn{1}{c|}{\textbf{F-Acc}} & \multicolumn{1}{c|}{\textbf{\begin{tabular}[c]{@{}c@{}}Q-Acc \\ (Activation)\end{tabular}}} & \multicolumn{1}{c|}{\textbf{\begin{tabular}[c]{@{}c@{}}Q-Acc \\ (Weights)\end{tabular}}} & \multicolumn{1}{c|}{\textbf{\begin{tabular}[c]{@{}c@{}}Q-Acc\\  (Both)\end{tabular}}} \\ \hline
\mbvone        & 70.558 & 70.48 & 70.394 & 70.418          \\ \hline
\mbvtwo        & 71.812 & 71.616 & 71.668       & 71.46           \\ \hline
\nasnet        & 74.376 & 74.068 & 74.352       & 73.888          \\ \hline
\vgg           & 70.956 & 70.834 & 70.946       & 70.81           \\ \hline
\inc           & 77.908 & 77.872 & 77.844       & 77.85           \\ \hline
\incres        & 80.284 & 80.154 & 80.32       & 80.14           \\ \hline
\res           & 75.106 & 75.072 & 75.06       & 75.018          \\ \hline
\eff           & 77.2   & 74.3 & 77.012       & 74.216          \\ \hline
\effrelu       & 77.65  & 77.1 & 77.568       & 77.092         \\ \hline
\dense         & 74.848 & 73.252 & 74.784       & 73.356          \\ \hline
\xecption      & 79.05  & 79.048 & 79.062       & 78.972          \\ \hline
\end{tabular}
\end{table}

\paragraph{Activation Quantization Analysis.}
In this analysis we evaluate the influence of the different methods used for quantizing the activations (without quantizing the weights). 
The analysis is performed with eleven different network architectures\footnote{\label{foot:eff}EfficientNet-B0 ReLU is a trained version of EfficientNet-B0 with ReLU activation function instead of swish}\footnote{ \url{https://keras.io/api/applications/}} on the ImageNet classification \cite{deng2009imagenet} task.
Table \ref{tab:activation_t} shows an accuracy comparison using four different threshold selection methods without applying any other of the activation quantization steps. 
NC indicates using the no-clipping threshold. Mean Square Error (MSE), Mean Average Error (MAE) and Kullback–Leibler (KL) are three different error measurements $d$ in Equation \ref{equ:err}.

\begin{table}[H]
\caption{ImageNet classification \cite{deng2009imagenet} accuracy with activations quantized using different threshold selection methods (weights are in floating point).
}
\label{tab:activation_t}
\begin{tabular}{|l|l|l|l|l|}
\hline
\multicolumn{1}{|c|}{\textbf{Network}}           &  \multicolumn{1}{|c|}{\textbf{NC}}       & \multicolumn{1}{|c|}{\textbf{MSE}}    & \multicolumn{1}{|c|}{\textbf{MAE}}    & \multicolumn{1}{|c|}{\textbf{KL}}   \\ \hline
\mbvone           & 70.406  & 70.434 & 60.218 & 70.418   \\ \hline
\mbvtwo           & 71.25 & 71.458 & 65.918  & 71.482   \\ \hline
%\nasnet           & 74.188 & 74.25  & 72.79  & 73.358   \\ \hline
\vgg              & 70.8 & 70.764 & 58.37 & 65.096   \\ \hline
%\inc              & 77.768 & 77.858 & 70.91  & 74.28   \\ \hline
%\incres           & 80.244 & 80.294 & 78.676 & 77.112   \\ \hline
\res              & 74.612 & 74.996 & 67.896 & 59.556  \\ \hline
%\eff              & 76.822 & 76.822 & 75.86  & 75.554   \\ \hline
%\effrelu          & 77.078 & 77.23  & 76.916 & 76.674   \\ \hline
%\dense            & 74.734 & 74.75  & 72.102 & 60.17   \\ \hline
%\xecption         & 79.006 & 79.01  & 77.47  & 75.374   \\ \hline
\end{tabular}
\centering
\end{table}

Table \ref{table:activation_ablation} shows the incremental accuracy influence on ImageNet classification \cite{deng2009imagenet} of the methods used by HPTQ for activation quantization (without quantizing weights).
Note that SNC is applied in all of the experiments in the table and its influence is studied separately below.
The table shows that all of the methods result in an improvement. Note that fine-tuning the z-score threshold $z_{th}$ per network may lead to further improvement.
% Specifically, equalization improves the accuracy in most cases even up to 2\% (\effrelu), where in the remaining cases there is only a negligible degradation. 
% Adding an MSE threshold results in accuracy improvement  for all cases, even by $60\%$ (\eff). 
% Adding a z-score with a threshold set to $z_{th}=24$ shows additional minor improvement in accuracy in most cases. 
% Fine-tuning the threshold $z_{th}$ may lead to further improvement.

\begin{table}[H]
\caption{
The accuracy influence of the different activation quantization methods used by HPTQ for ImageNet classification \cite{deng2009imagenet} when keeping all weights in floating point.
Baseline is quantization with no-clipping thresholds, \textit{+Eq.} means adding max channel equalization,
\textit{+MSE Th.} means replacing the no-clipping thresholds with MSE
and \textit{+z-score} means applying z-score outlier removal.
}
\label{table:activation_ablation}
\centering
\begin{tabular}{|l|l|l|l|l|l|}
\hline
\multicolumn{1}{|c|}{\textbf{Network Name}}            & \multicolumn{1}{|c|}{\textbf{Baseline}} & \multicolumn{1}{|c|}{\textbf{+Eq.}} & \multicolumn{1}{|c|}{\textbf{+MSE Th.}} & \multicolumn{1}{|c|}{\textbf{+z-score}} \\ \hline
\mbvone                 & 70.406          & 70.418                 &70.48                &    70.48            \\ \hline
\mbvtwo                 & 71.25         &   71.34               &      71.528            &      71.616           \\ \hline
\nasnet                 &   18.572       &     18.484             &  73.486              &   74.068              \\ \hline
\vgg                    &   70.8       &     70.696            &  70.888              &   70.834              \\ \hline
\inc                    &  77.658       &     77.646            &  77.832              &   77.872             \\ \hline
\incres                 & 49.132        & 49.238                &   80.014              &        80.154         \\ \hline
\res                    & 74.612         & 74.654                &   75.086              &        75.072         \\ \hline
\eff                   &   13.562        &     13.736            &  74.096               &   74.3               \\ \hline
\effrelu                &   74.298        &     76.298             &  76.956               &   77.1               \\ \hline
\dense                  &  56.08       &     55.916            &  73.28              &   73.252           \\ \hline
\xecption               &   48.718       &     48.784             &  78.87               &   79.048               \\ \hline
\end{tabular}
\end{table}

%\paragraph{Shift Negative Correction (SNC) Analysis.}
% To evaluate the SNC effect, we trained several versions of MobileNetV1, each with a different Non-Linear function, with a full flow of activation quantization. 
% For more details about the training process see Appendix \ref{subsec:mbv1_training}. 
Table \ref{table:snc} shows the accuracy improvement achieved by applying Shift Negative Correction (SNC). Specifically, the table compares the performance of several versions of MobileNetV1, each with different non-linear functions, with a full flow of activation quantization.

\begin{table}[H]
\centering
\caption{ImageNet classification accuracy \cite{deng2009imagenet} using HPTQ with and without SNC of \mbvone trained with different non-linear functions.}
\label{table:snc}
\begin{tabular}{|l|l|l|l|l|l|l|}
\hline
            & \multicolumn{1}{|c|}{\textbf{Swish}}              & \multicolumn{1}{|c|}{\textbf{\begin{tabular}[c]{@{}c@{}}Leaky ReLU\\ ($\alpha=0.1$)\end{tabular}}}  & \multicolumn{1}{|c|}{\textbf{PReLU}}  & \multicolumn{1}{|c|}{\textbf{SELU}}   \\ \hline
Float     & 73.522                & 72.866                                 & 73.114 & 72.032 \\ \hline
Without SNC & 60.98 &71.966       & 72.548 & 69.726 \\  \hline
With SNC  & 71.146           & 72.588                                                               & 72.548 & 70.902 \\ \hline
\end{tabular}
\end{table}

% \begin{table}[H]
% \centering
% \caption{HPTQ accuracy for ImageNet classification \cite{deng2009imagenet} with/without SNC on \mbvone trained with different non-linear functions.}
% \label{table:snc}
% \begin{tabular}{|l|l|l|l|l|l|l|}
% \hline
%             & \multicolumn{1}{|c|}{\textbf{Swish}} & \multicolumn{1}{|c|}{\textbf{ReLU6}}                   & \multicolumn{1}{|c|}{\textbf{ReLU}}                    & \multicolumn{1}{|c|}{\textbf{\begin{tabular}[c]{@{}c@{}}Leaky ReLU\\ ($\alpha=0.1$)\end{tabular}}}  & \multicolumn{1}{|c|}{\textbf{PReLU}}  & \multicolumn{1}{|c|}{\textbf{SELU}}   \\ \hline
% Float     & 73.522  & 72.356                  & 72.884                  & 72.866                                                               & 73.114 & 72.032 \\ \hline
% Without SNC & 60.98 & \multirow{2}{*}{72.246} & \multirow{2}{*}{72.662} &71.966                                                                & 72.548 & 69.726 \\ \cline{1-2} \cline{5-7} 
% With SNC  & 71.146  &                         &                         & 72.588                                                               & 72.548 & 70.902 \\ \hline
% \end{tabular}
% \end{table}

\paragraph{Weight Quantization Analysis.} 
In this analysis we evaluate the influence of the different methods used for quantizing weights (without quantizing activations). 
The analysis is performed with eleven different network architectures\footnote{\ref{foot:eff}EfficientNet-B0 ReLU is a trained version of EfficientNet-B0 with ReLU activation function instead of swish}\footnote{ \url{https://keras.io/api/applications/}} on the ImageNet classification \cite{deng2009imagenet} task.

Table \ref{tab:weights_t} shows an accuracy comparison of each quantized network using four different threshold selection methods (without applying bias correction).
NC indicates using the no-clipping threshold. Mean Square Error (MSE), Mean Average Error (MAE) and Kullback–Leibler (KL) are three different error measurements $d$ in Equation \ref{equ:err}.
Similarly to the results for activation quantization in Table \ref{tab:activation_t}, the MSE error measurement achieves the best results.

%Following this observation, we have decided to only use MSE as a threshold metric for weight quantization in the following experiments.

\begin{table}[H]
\caption{ImageNet classification \cite{deng2009imagenet} accuracy with weights quantized using different threshold selection methods (activations are in floating point).}
\label{tab:weights_t}
\begin{tabular}{|l|l|l|l|l|}
\hline
\multicolumn{1}{|c|}{\textbf{Network}}           &  \multicolumn{1}{|c|}{\textbf{NC}}       & \multicolumn{1}{|c|}{\textbf{MSE}}    & \multicolumn{1}{|c|}{\textbf{MAE}}    & \multicolumn{1}{|c|}{\textbf{KL}}   \\ \hline
\mbvone           & 68.75  & 68.756 & 64.242 & 64.968   \\ \hline
\mbvtwo           & 69.562 & 69.758 & 67.57  & 62.394   \\ \hline
\nasnet           & 74.188 & 74.232    & 72.79  & 73.358   \\ \hline
\vgg              & 70.944 & 70.94  & 67.486 & 70.472   \\ \hline
\inc              & 77.768 & 77.82& 70.91  & 74.28   \\ \hline
\incres           & 80.244 & 80.276  & 78.676 & 77.112   \\ \hline
\res              & 75.068 & 75.11 & 72.352 & 73.418   \\ \hline
\eff              & 76.822 & 76.822 & 75.86  & 75.554   \\ \hline
\effrelu          & 77.078 & 77.218  & 76.916 & 76.674   \\ \hline
\dense            & 74.734 & 74.736  & 72.102 & 60.17   \\ \hline
\xecption         & 79.006 & 79.006  & 77.47  & 75.374   \\ \hline
\end{tabular}
\centering
\end{table}

Table \ref{table:weights_ablation} shows the incremental accuracy influence of the two methods (per channel quantization and bias correction) used in HPTQ for weight quantization (without quantizing activations) on the ImageNet classification task \cite{deng2009imagenet}.
This table shows that both of our methods result in improvement.
% In the results under $MSE th.$, weights are quantized per layer and their thresholds are selected using MSE as the error measurement.
% For some networks (\mbvone and \mbvtwo) it results in a large accuracy improvement, but for the other networks, it either results in a small improvement (\nasnet) or even in an accuracy degradation (\res). 

\begin{table}[H]
\caption{
The incremental influence of applying per-channel threshold selection (\textit{Per ch.}) and bias correction (\textit{Bias corr.}) on ImageNet \cite{deng2009imagenet} classification accuracy.
Baseline means quantization with MSE threshold applied per tensor.
}
\label{table:weights_ablation}
\centering
\begin{tabular}{|l|l|l|l|l|l|}
\hline
\multicolumn{1}{|c|}{\textbf{Network}}             & \multicolumn{1}{|c|}{\textbf{Baseline}} & \multicolumn{1}{|c|}{\textbf{Per ch.}} & \multicolumn{1}{|c|}{\textbf{+Bias corr.}} \\ \hline
\mbvone                    & 0.966                &  68.756                 &    70.394             \\ \hline
\mbvtwo                    &   0.398              &  69.758            &      71.668           \\ \hline
\nasnet                     &     73.494          &  74.232               &  74.352               \\ \hline
\vgg                      &     70.814            &  70.94               &  70.946                \\ \hline
\inc                      &     76.42             &  77.82               &  77.844               \\ \hline
\incres                  &     80.066             &  80.276               &  80.32               \\ \hline
\res                      & 74.718                &  75.11              &        75.06          \\ \hline
\eff                       &     2.524            &  76.822             &  77.012            \\ \hline
\effrelu                   &     0.682            &  77.218             &  77.568            \\ \hline
\dense                  &     72.986              &  74.736               &  74.784             \\ \hline
\xecption                &     78.786             &  79.006               &  79.062               \\ \hline
\end{tabular}
\end{table}